\documentclass[conference]{IEEEtran}

\ifCLASSINFOpdf
\else
\fi

\usepackage{multirow}
\usepackage{amsmath}
\usepackage{hyperref}
\usepackage{url}

\usepackage[utf8]{inputenc}
\usepackage[english]{babel}
\usepackage{biblatex}
\usepackage{setspace}
\usepackage{threeparttable}
\usepackage{footnote}

\usepackage[all,defaultlines=2]{nowidow}

\usepackage{tablefootnote}

\newcommand{\half}{\frac{1}{2}}
\usepackage{verbatim}

\usepackage[table]{xcolor}
\usepackage{eso-pic}
\usepackage[disable]{todonotes}
\definecolor{Gray}{gray}{0.925} 
\definecolor{Preprint}{rgb}{.63,.79,.95}

\newcommand{\preprintBanner}{
\AddToShipoutPictureFG*{\put(\LenToUnit{0.5\paperwidth},\LenToUnit{0.925\paperheight}){\makebox[0pt][c]{
\renewcommand{\arraystretch}{1.5}
\setlength{\tabcolsep}{18pt}
\rowcolors{1}{Preprint}{Gray}
\begin{tabular}{|p{0.55\paperwidth}|} \hline
\textbf{How to cite:} \\ \hline
\footnotesize
I. Venugopal, J. Töllich, M. Fairbank, A. Scherp, ``A Comparison of Deep-Learning Methods for Analysing and Predicting Business Processes'' in \textit{Proceedings of International Joint Conference on Neural Networks, {IJCNN}}, IEEE Press, July 18-22, 2021.\\ \hline
\end{tabular}}}}
}

\begin{document}

\title{\vspace{2cm}A Comparison of Deep-Learning Methods for Analysing and Predicting Business Processes}

\author{\IEEEauthorblockN{Ishwar Venugopal}
\IEEEauthorblockA{University of Essex, UK\\
iv19023@essex.ac.uk}
\and
\IEEEauthorblockN{Jessica Töllich}
\IEEEauthorblockA{Ulm University, Germany\\
jessica.toellich@uni-ulm.de}
\and
\IEEEauthorblockN{Michael Fairbank}
\IEEEauthorblockA{University of Essex, UK
\\m.fairbank@essex.ac.uk}
\and
\IEEEauthorblockN{Ansgar Scherp}
\IEEEauthorblockA{Ulm University, Germany\\
ansgar.scherp@uni-ulm.de}}

\maketitle

\preprintBanner

\begin{abstract}
Deep-learning models such as Convolutional Neural Networks (CNN) and Long Short-Term Memory (LSTM) have been successfully used for process-mining tasks. They have achieved better performance for different predictive tasks than traditional approaches. We extend the existing body of research by testing four different variants of Graph Neural Networks (GNN) and a fully connected Multi-layer Perceptron (MLP)  with dropout for the tasks of predicting the nature and timestamp of the next process activity. In contrast to existing studies, we evaluate our models’ performance at different stages of a process, determined by quartiles of the number of events and normalized quarters of the case duration. This provides new insights into the performance of a prediction model, as they behave differently at different stages of a business-process. Interestingly, our experiments show that the simple MLP often outperforms more sophisticated deep-learning models in both prediction tasks. We argue that care needs to be taken when applying automated process-prediction techniques at different stages of a process. We further argue that researchers should reflect their results with strong baselines methods like MLPs. 
\end{abstract}

\IEEEpeerreviewmaketitle

\section{Introduction}

Most businesses thrive on the effective use of event logs and process records. The ability to predict the nature of an unseen event in a business-process can have very useful applications \cite{breuker2016comprehensible}. This can help in more efficient customer service, and facilitate in developing an improved work-plan for companies. 
The domain of process mining deals with combining a wide range of classical model-based predictive techniques along with traditional data-analysis techniques~\cite{van2016data}.  
A process can be a representation of any set of activities that take place in a business enterprise; for example, the procedure for obtaining a financial document, the steps involved in a complaint-registering system, etc. 

Business-process mining, in general, deals with the analysis of the sequence of events produced during the execution of such processes~\cite{castellanos2004comprehensive,maggi2014predictive,marquez2017predictive}. 
Even though the classical approach of depicting event logs is with the help of process graphs \cite{agrawal1998mining,van2003workflow}, 
Pasquadibisceglie et al.~\cite{pasquadibisceglie2019using},
Tax et al.~\cite{tax2017predictive}, Taymouri et al.~\cite{taymouri2020predictive}, and others have recently applied deep-learning techniques like Convolutional Neural Networks (CNN), Long Short-Term Memory (LSTM) networks, and Generative Adversarial Nets (GANs) for the task of predictive process mining.
The deep-learning based models obtained results that outperformed traditional models. 

Inspired from these works and taking into consideration the graph nature of processes, we aim to model event logs as graph structures and apply different Graph Neural Network (GNN) models on such data structures. 
GNNs have shown superior results for the vertex-classification task ~\cite{DBLP:conf/iclr/KipfW17}, link-prediction task~\cite{zhang2018link}, and recommender systems~\cite{ying2018graph}.
In this work, we use a new representation for the event-log data and investigated the performance of different variants of a Graph Convolutional Network (GCN)~\cite{DBLP:conf/iclr/KipfW17} as a successful example of GNNs.
We compare the GCN model among others with the CNN and LSTM models along with a Multi-Layer Perceptron (MLP) and classical process-mining techniques~\cite{breuker2016comprehensible,van2011time}.

In contrast to the existing body of research~\cite{pasquadibisceglie2019using,evermann2016deep,camargo2019learning,tax2017predictive,taymouri2020predictive,lin2019mm}, we analyze how the performance of the models for business-process prediction depend on the stage of a process.
The results show that the next activity type and timestamp prediction depend a lot on the model and also on whether an early, mid, or late stage of the process is considered.
Furthermore, we observe from our experiments that MLP is a strong baseline and in many cases outperforms more advanced neural networks like LSTMs, GCNs, and CNNs. 
The MLP model achieves a maximum of 82\% accuracy in predicting the next event type, and a minimum mean absolute error of 1.3229 days for predicting the timestamp of the next event.

Below, we discuss related works in business-process mining. 
Section~\ref{sec:exp_app} introduces our experimental apparatus, datasets and pre-processing, as well as our GCN-based models.
Sections~\ref{sec:results} and ~\ref{sec:results_comparison} highlights the major results from the experiments, followed by a discussion in Section~\ref{sec:discussions}, before we conclude.

\section{Related Works}
\label{sec:rel_works}

Business-process mining deals with several prediction tasks like predicting the next activity type~\cite{becker2014designing,tax2017predictive,pasquadibisceglie2019using,evermann2016deep,breuker2016comprehensible}, the timestamp of the next event in the process~\cite{tax2017predictive,van2011time}, the overall outcome of a given process~\cite{taylor2017customer}, or the time remaining until the completion of a given process instance \cite{rogge2013prediction}. 
There is a huge body of algorithms for these process-mining tasks~\cite{breuker2016comprehensible,van2011time}.
In the context of this work, we focus on the first two aspects of the aforementioned list of predictive tasks, namely, the task of predicting the nature and timestamp of the next event in a given process.
We reconsider the results from the classical methods and compare them with latest developments on business-process mining using deep learning. 

There has been a recent shift towards deep-learning models for the task of predictive business-process monitoring. 
Tax et al.~\cite{tax2017predictive} proposed to use a Recurrent Neural Network (RNN) architecture with Long Short-Term Memory (LSTM) for the task of predicting the next activity and timestamp, the remaining cycle time, and the sequence of remaining events in a case.
Their model was able to model the temporal properties of the data and improve on the results obtained from traditional process-mining techniques. 
The main motivation for using an LSTM model was to obtain results that were consistent for a range of tasks and datasets. 
The LSTM architecture of Tax et al. could also be extended to the task of predicting the case outcome. 
Camargo et al.~\cite{camargo2019learning} and Lin et al.~\cite{lin2019mm} both use LSTM models, too.
The first one to predict the next event including timestamp and the associated resource pool, the latter to predict the next event, including its attributes.
Evermann et al. \cite{evermann2016deep} also used RNN for the task of predicting the next event on two real-life datasets. Their system architecture involved two hidden RNN layers using basic LSTM cells. 

Pasquadibisceglie et al. \cite{pasquadibisceglie2019using} used Convolutional Neural Networks (CNN) for the task of predictive process analytics. 
An image-like data engineering approach was used to model the event logs and obtain results from benchmark datasets. 
In order to adapt a CNN for process-mining tasks, a novel technique of transforming temporal data into a spatial structure similar to images was introduced. 
The CNN results improve over the accuracy scores obtained by Tax et al.'s LSTM~\cite{tax2017predictive} for the task of predicting the next event. 

Scarselli et al. \cite{scarselli2008graph} introduced Graph Neural Networks (GNNs) as a new deep-learning technique that could efficiently perform feature extraction. 
Especially in the last year, GNNs have gained widespread attention and use in different domains.
Wu et al. \cite{wu2020comprehensive} provided a comprehensive survey of GNNs.
They categorize the different GNN architectures into Graph  Convolutional Networks  (GCN, or also called: ConvGNN), Spatio-temporal Graph Neural Networks (STGNNs), Recurrent Graph Neural Networks (RecGNN), and Graph Autoencoders (GAEs).
Esser et al. \cite{a3a7ca89d76a435ca35751963ce60f18} discussed the advantages of using graph structures to model event logs. Performing process-mining tasks by modelling the relationships between events and case instances as process graphs has been a widely accepted approach \cite{maruster2002process,van2007business}. 

Recently, Taymouri et al. \cite{taymouri2020predictive} have used Generative Adversarial Nets (GANs) for predicting the next activity and its timestamp. 
In a minmax game of discriminator and generator, both consisting of RNNs in a LSTM architecture and feedforward neural networks, a prediction is made of the next step, including event type and event-timestamp prediction.
Taymouri et al. used different models each trained over a specific length of sub-sequences of processes, modeled by the parameter $k$.
For example, a value of $k=20$ means that sub-sequences of length $20$ are used for training, and testing would be applied on process steps $21$, $22$, $23$, and following until the end of the process.

Other works used features from unstructured data like texts in deep-learning architectures to improve the process-prediction task. 
Ding et al.~\cite{ding2015deep} demonstrate how a deep-learning model using events extracted from texts improves predictions in the stock markets domain. 
For business-process modelling, Teinemaa et al.~\cite{teinemaa2016predictive} improve the performance of predictive business models by using text-mining techniques on the unstructured data present in event logs.

In this work, we aim to combine traditional process mining from event graphs along with deep-learning techniques like GCNs to achieve a better performance in predictive business-process monitoring. 
We evaluate each of the model variants at different stages of a process, determined by quartiles of the number of events in a case
and normalized quarters computed over the case durations. 
This would provide a more detailed understanding of the models' performance.

\section{Experimental Apparatus}
\label{sec:exp_app}

We introduce the datasets used in this work and the methodology adopted for representing the feature vectors corresponding to each row in the dataset. Following this, a mathematical formulation of graphs and the specific case of process graphs is provided, which lays the foundation to understand a Graph Convolutional Network. 
We conclude with a description of the procedure and metrics adopted for this work. 

\subsection{Datasets}
\label{sec:datasets}

We use two well-known benchmark event-log datasets, namely Helpdesk and BPI12 (W).
These two representative datasets have been chosen as they are used by the models we want to compare with, namely the CNN by Pasquadibisceglie et al.~\cite{pasquadibisceglie2019using}, LSTMs from Camargo et al.~\cite{camargo2019learning} and Tax et al.~\cite{tax2017predictive}, and the GAN from Taymouri et al.~\cite{taymouri2020predictive}.
Thus, the datasets best possible serve the purpose to compare the different Deep-Learning architectures.
All datasets are characterised by three columns: ``Case ID'' (the process-case identifier), ``Activity ID'' (the event-type identifier), and the ``Complete Timestamp'' denoting the time at which a particular event took place. 
Table~\ref{table:data-analysis} shows an overview of the datasets.

\begin{table}[!h]
\centering
\caption{Overview of the datasets used}
\label{table:data-analysis}
\begin{tabular}{l|rr}
\multicolumn{1}{c|}{\multirow{2}{*}{\textbf{Attribute}}} & \multicolumn{2}{c}{\textbf{Dataset}}                           \\ \cline{2-3} 
\multicolumn{1}{c|}{}                                    & \multicolumn{1}{c|}{Helpdesk} & \multicolumn{1}{c}{BPI12(W)} \\ \hline
No. of events             & \multicolumn{1}{r|}{13,710}    & 72,413                          \\
No. of process cases         & \multicolumn{1}{r|}{3,804}     & 9,658                           \\
No. of activity types     & \multicolumn{1}{r|}{9}        & 6                              \\
Avg. case duration (sec.) & \multicolumn{1}{r|}{22,475} & 1,364                        \\
Avg. no. of events per case   & \multicolumn{1}{r|}{3.604}    & 7.498                         
\end{tabular}
\end{table}

\paragraph{Helpdesk dataset}
This dataset presents event logs obtained at the helpdesk of an Italian software company.\footnote{\url{https://data.mendeley.com/datasets/39bp3vv62t/1}}
The events in the log correspond to the activities associated with different process instances of a ticket management scenario. 
It is a database of 13,710 events related to 3,804 different process instances. 
There are 9 activity types, i.\,e., classes in the dataset.
Each process contains events from a list of nine unique activities involved in the process. 
A typical process instance spans events from inserting a new ticket, until it is closed or resolved. 
Table \ref{table:data-analysis} shows the average case duration and the number of activities per case.

\paragraph{BPIC'12 (Sub-process W) dataset}
The Business-Process Intelligence Challenge (BPIC'12) dataset\footnote{\url{https://www.win.tue.nl/bpi/doku.php?id=2012:challenge&redirect=1id=2012/challenge}} contains event logs of a process consisting of three sub-processes, which in itself is a relatively large dataset. 
As described in \cite{tax2017predictive} and \cite{pasquadibisceglie2019using}, only completed events that are executed manually are taken into consideration for predictive analysis. 
This dataset, called BPI12 (W), includes 72,413 events from 9,658 process instances. 
Each event in a process is one among 6 activity types involved in a process instance, i.\,e., a process case. 
The activities denote the steps involved in the application procedure for financial needs, like personal loans and overdrafts.

\subsection{Graphs and Graph Convolutional Layer}
\label{subsec:GCN}
A graph can be represented as $G = (V,E)$, where \textit{V} is the set of vertices and \textit{E} denotes the edges present between the vertices~\cite{wu2020comprehensive}.
An edge between vertex \textit{i} and vertex \textit{j} is denoted as  \textit{e$_{ij}$} $\in$ \textit{E}.  A graph can be either directed or undirected, depending on the nature of interaction between the vertices. 
In addition, a graph may be characterized by vertex attributes or edge attributes, which in simple terms are feature vectors associated with that particular vertex or edge. 
The adjacency matrix of a graph is an \textit{$n \times n$} matrix with \textit{A$_{ij}$} = 1  if e$_{ij}$ $\in$ \textit{E} and \textit{A$_{ij}$} = 0  if e$_{ij}$ $\notin$ \textit{E}, where \textit{n} is the number of vertices in the graph. 
A degree matrix is a diagonal matrix which stores the degree of each vertex, which numerically corresponds to the number of edges that the node is attached to. 

A GCN layer operates by calculating a hidden embedding vector for each node of the graph.  It calculates this hidden vector by combining each node's feature-vector with the adjacency matrix for the graph, by the equation (Kipf and Welling~\cite{DBLP:conf/iclr/KipfW17}):
\begin{equation}
    f(X,A,W) = \sigma (D^{-1}AXW), \label{equation:gcn_kipfwelling}
\end{equation}
where \textit{X} is the input-feature matrix containing the feature vector for each of the vertices, \textit{A} is the adjacency matrix of the graph, \textit{D} is the degree matrix, \textit{W} is a learnable weight matrix, and $\sigma$ is the activation function of that layer. 

In \eqref{equation:gcn_kipfwelling}, the product $D^{-1}A$ represents an attempt to normalize the adjacency matrix. However, as matrix multiplication is non-commutative, an alternative symmetric normalisation is preferred \cite{DBLP:conf/iclr/KipfW17}, changing the GCN layer's operation to:
\begin{equation}
    \label{equation:d_normal}
    f(X,A,W) = \sigma (D^{-\half}AD^{-\half}XW)
\end{equation}

Note, all the models used in this work, the GCN layer calculations are done as described in \eqref{equation:d_normal}. 
For the model variants described in Section \ref{subsec:model_variants} involving the Laplacian matrix, the adjacency matrix ($A$) in \eqref{equation:d_normal} is replaced by the corresponding unnormalized Laplacian.

\subsection{Data Pre-processing}
\label{section:feat_vec}

The timestamp corresponding to each event in the dataset can be used to derive a feature-vector representation for each row in the data. The approach introduced in \cite{tax2017predictive} has been used to initially get a feature vector consisting of the following four elements: 1. The time since previous event in the case. 2.~The time since the case started. 3. The time since midnight. 4. The day of the week for the event.
All four values are treated as real-valued durations. 
This results in a 4-element feature vector for every row in the dataset. 
The drawback in this kind of a representation is that it treats each event in a case independently. In order to overcome this drawback, it was necessary for the feature vector of every event to have a history of other events that had already occurred for that particular Case ID. Hence, a new comprehensive feature vector representation was introduced. 

In this work, each entry in a dataset is assigned a matrix representation 
(\emph{X}) whose dimensions depend on the dataset which is considered. 
The number of \emph{rows} in
\emph{X} can be obtained by identifying the unique entries in the `Activity ID' column, i.\,e., the unique activity types as shown in Table~\ref{table:data-analysis}, or can be visually identified as the number of vertices in the process graphs for each of the datasets (Figure \ref{fig:dfg}). 
Let us denote this value by `\textit{num$\_$of$\_$nodes}' for ease of representation. As it can be observed from Table \ref{table:data-analysis}, \textit{num$\_$of$\_$nodes} is 9 for the Helpdesk dataset and 6 for the BPI'12 (W) dataset. 
The number of \emph{columns} in
\emph{X} corresponds to the length of the initial feature vector, i.\,e., 4. 
This would result in a matrix of size `\textit{$num\_of\_nodes \times 4$}' for each data entry. 

The matrix \emph{X} is first initialized with zeroes. Each row index of $X$ stores the 4-element long feature vector corresponding to the most-recent Activity ID denoted by that particular row index, for the current case ID. 
For example, the first row stores the 4-element long feature vector for the event with Activity ID equal to 1,
and so on. 
One approximation that we have used in this step is that if an event corresponding to a particular Activity ID has occurred more than once in a case, we use the feature vector for only the most-recent occurrence of that event.
In scenarios where events with a particular Activity ID have not occurred yet in a given case, the feature matrix will hence just store a vector with zeroes corresponding to that Activity ID. 
This method of representation gives each row of the Helpdesk dataset a $9 \times 4$ matrix, and each row of the BPI'12 (W) dataset a $6 \times 4$ matrix. 
The motivation behind choosing such a representation is to facilitate the computation involved in a Graph Convolutional Layer, as explained in Section \ref{subsec:GCN}. 
For a given row, the Activity ID of the next event and the time since current event are taken as the target labels for the event-predictor and the time-predictor, respectively.

\subsection{Process Graphs as Input to GCNs}

\begin{figure} [!h]
    \centering
    \includegraphics[width=0.49\linewidth]{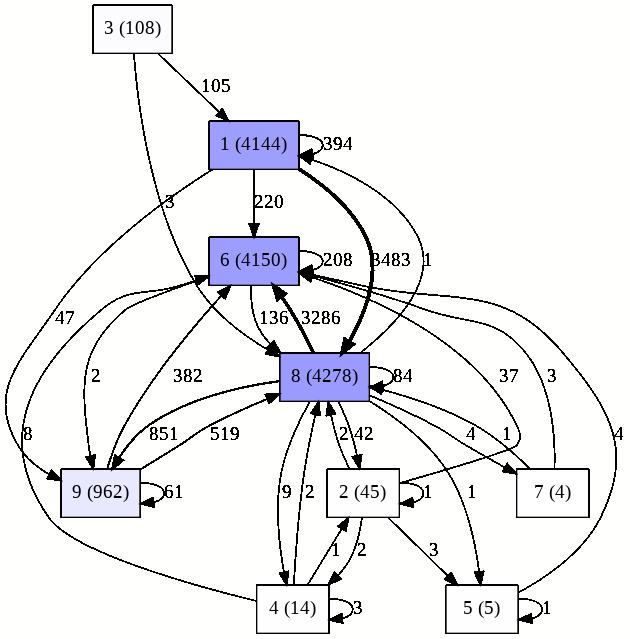}
    \includegraphics[width=0.49\linewidth]{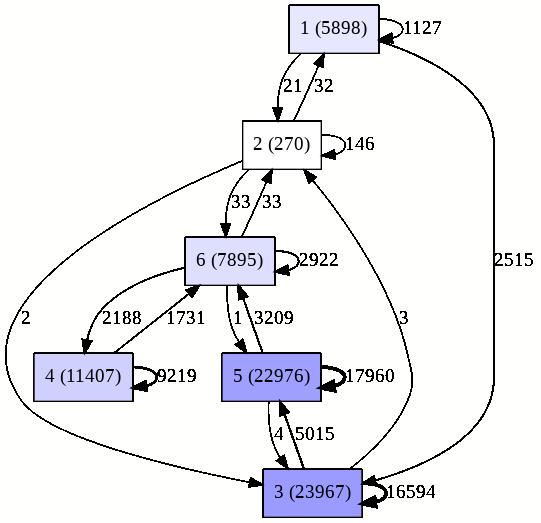}
    \caption{Directly-follows graphs generated for the Helpdesk dataset (left) and BPI'12 (W) dataset (right) using PM4Py. The vertices represent the unique Activity IDs (i.\,e., activity types) along with their frequencies denoted in brackets. The numbers on the directed edges denote the frequency of directly-follows relations.} 
    \label{fig:dfg}
\end{figure}

\begin{figure*} [!h]
    \centering
    \includegraphics[width=1\linewidth]{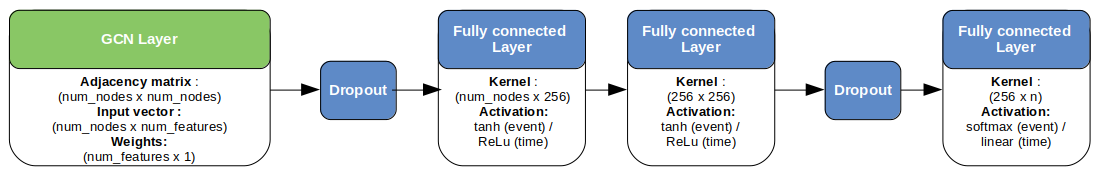}
    \caption{Graph Convolutional Network architecture for the event type and timestamp predictor. The value for \textit{n} in the last layer denotes the number of classes for the event predictor and 1 for the time predictor.
    }
    \label{fig:system_arch}
\end{figure*}

Process discovery from event logs can be achieved using different traditional process-mining techniques. In this work, we have used an inductive mining approach with Directly-Follows Graphs (DFGs) to represent the processes extracted from each of the datasets. The choice is motivated by the simplicity and efficiency with which the entire data can be represented in the form of a graph. 

A Directly-Follows Graph for an Event Log \textit{L} is denoted as \cite{van2016data}:
$G(L) = (A_L,\mapsto _L,A_L^{start},A_L^{end})$,
where A$_L$ is the set of activities in \textit{L} with A$_L^{start}$ and A$_L^{end}$ denoting the set of start and end activities, respectively. $\mapsto _L$ denotes the directly-follows operation, which exists between two events if and only if there is a case instance in which the source event is followed by the other target event. 
The vertices in the graph represent the unique activities present in the event log, and the directed edges of the graph exist if there is a directly-follows relation between the vertices. 
The number of directly-follows relations that exist between two vertices is denoted by a weight for the corresponding edge.  

Berti et al.~\cite{DBLP:journals/corr/abs-1905-06169} presented a process-mining tool for Python called PM4Py. 
The Directly-Follows Graphs for both the datasets (considering all the events/rows) were visualised using the PM4Py package as shown in Figure \ref{fig:dfg}. 
Consider the following binary adjacency matrix for the process graph generated, as example, from the BPI'12 (W) dataset:

{\small
\[
B_{BPI'12(W)} = 
\begin{bmatrix}
  1 & 1 & 1 & 0 & 0 & 0\\
  1 & 1 & 1 & 0 & 0 & 1\\
  0 & 1 & 1 & 0 & 1 & 0\\
  0 & 0 & 0 & 1 & 0 & 1\\
  0 & 0 & 1 & 0 & 1 & 1\\
  0 & 1 & 0 & 1 & 1 & 1\\
\end{bmatrix}
\]
}

This \textit{$6 \times 6$} matrix needs to be normalized as per Equation \eqref{equation:d_normal} 
to be used in a GCN. 
The elements of the diagonal degree matrix can be numerically computed as a row-wise sum from the above matrix. 
The dimensions of the normalized matrix (\textit{$6 \times 6$}) and the dimension of \emph{X} (\textit{$6 \times 4$} for BPI'12 (W) dataset) makes it compatible for matrix multiplication in the GCN layer. 
In general, the normalized adjacency matrix will have dimensions \textit{$num\_of\_nodes \times num\_of\_nodes$} and \emph{X} the dimensions \textit{$num\_of\_nodes \times 4$}.

\subsection{Procedure}

The network depicted in Figure \ref{fig:system_arch} shows the architecture for the GCN model that learns the next Activity ID and the timestamp of the next activity. The overall structure that was constructed for this work mainly focuses on a Graph Convolutional Layer followed by a sequential layer consisting of three fully-connected layers with Dropout (present after the GCN layer and before the last fully-connected layer). 
The weight matrix (W) in the GCN layer is of size $4 \times 1$.
The Event Predictor Network has \textit{tanh} activation for the first two fully connected layers and softmax activation at the last layer. 
Cross-entropy loss is used during training. 
The Timestamp Predictor Network on the other hand consists of \textit{ReLU} activation for the first two layers and a linear activation function at the last layer. 
The training process uses the Mean Absolute Error as the loss function. 
An Adam optimizer \cite{DBLP:journals/corr/KingmaB14} is used for  the training processes for all variants. 
In line with the training procedure of prior studies~\cite{tax2017predictive,pasquadibisceglie2019using}, each of the datasets is divided into train (2/3) and test sets (1/3). 
We use 20\% of the training as validation set during the training process. 
The validation set is randomly sampled from the training set in each of the five experimental runs.
Note, the chronological nature of the datasets have been preserved during the train-test splitting. 
One row is taken at a time during training resulting in a mini-batch size of 1. 
The final results after the evaluation on the test set are reported as an average measure of 5 runs.

\begin{table*}[t!]
\centering
\caption{Accuracy for next-event prediction at different stages of a process (indicated by quartiles based on the number of events and quarters based on normalising the case duration). Standard deviations (SD) have been omitted as they are very low ($<0.06$). 
}
\label{table:accuracy_results}
\begin{tabular}{c|c|cccc|cccc|c}
\multirow{3}{*}{\textbf{Dataset}}                                      & \multirow{3}{*}{\textbf{Model}} & \multicolumn{8}{c|}{\textbf{Accuracy for Event Prediction}}                                                                                                                                                                                   & \multirow{3}{*}{\textbf{\begin{tabular}[c]{@{}c@{}}Overall\\ accuracy\end{tabular}}} \\ \cline{3-10}
                                                                      &                                 & \multicolumn{4}{c|}{\textbf{Quartiles  based  on  Events}}                                                            & \multicolumn{4}{c|}{\textbf{Quarters based on Duration}}                                                              &                                                                                      \\ \cline{3-10}
                                                                      &                                 & \multicolumn{1}{c|}{\textbf{1}} & \multicolumn{1}{c|}{\textbf{2}} & \multicolumn{1}{c|}{\textbf{3}} & \textbf{4}      & \multicolumn{1}{c|}{\textbf{1}} & \multicolumn{1}{c|}{\textbf{2}} & \multicolumn{1}{c|}{\textbf{3}} & \textbf{4}      &                                                                                      \\ \hline
\multirow{5}{*}{Helpdesk}                                              & GCN$_W$                            & 0.7288                          & 0.6888                          & 0.7634                          & 0.9419          & 0.7499                          & 0.5508                          & 0.5940                          & 0.8951          & 0.7954                                                                               \\
                                                                      & GCN$_B$                            & 0.7266                          & 0.6778                          & 0.7475                          & 0.8973          & 0.7418                          & 0.5410                          & 0.5590                          & 0.8561          & 0.7731                                                                               \\
                                                                      & GCN$_{LB}$                           & 0.7270                          & 0.6837                          & 0.7729                          & 0.9108          & 0.7523                          & 0.5492                          & 0.5819                          & 0.8722          & 0.7863                                                                               \\
                                                                      & GCN$_{LW}$                           & 0.6681                          & 0.6922                          & 0.7665                          & 0.9167          & 0.7389                          & 0.5508                          & 0.5723                          & 0.8803          & 0.7830                                                                               \\
                                                                      & MLP                             & \textbf{0.7297}                 & \textbf{0.7031}                 & \textbf{0.8110}                 & \textbf{0.9642} & \textbf{0.7677}                 & \textbf{0.6082}                 & \textbf{0.6446}                 & \textbf{0.9212} & \textbf{0.8201}                                                                      \\ \hline
\multirow{5}{*}{\begin{tabular}[c]{@{}c@{}}BPI'12 \\ (W)\end{tabular}} & GCN$_W$                            & 0.6964                          & 0.7397                          & 0.8011                          & 0.4303          & 0.7247                          & 0.8802                          & 0.7869                          & 0.4493          & 0.6484                                                                               \\
                                                                      & GCN$_B$                            & 0.7329                          & 0.7487                          & 0.8039                          & 0.3936          & 0.7424                          & 0.8819                          & 0.7933                          & 0.4251          & 0.6473                                                                               \\
                                                                      & GCN$_{LB}$                           & \textbf{0.7381}                 & \textbf{0.7587}                 & \textbf{0.8111}                 & 0.4077          & \textbf{0.7579}                 & \textbf{0.8961}                 & 0.7883                          & 0.4329          & \textbf{0.6569}                                                                      \\
                                                                      & GCN$_{LW}$                           & 0.7366                          & 0.7542                          & 0.8050                          & 0.4028          & 0.7552                          & 0.8827                          & 0.7882                          & 0.4279          & 0.6525                                                                               \\
                                                                      & MLP                             & 0.6554                          & 0.7369                          & 0.8058                          & \textbf{0.4792} & 0.7006                          & 0.8818                          & \textbf{0.8001}                 & \textbf{0.4888} & 0.6559                                                                              
\end{tabular}
\end{table*}
\begin{table*}[t!]
\centering
\caption{MAE values (in days) for predicting the timestamp of the next-event at different stages of a process (indicated by quartiles based on the number of events and quarters based on normalising the case duration). 
SDs omitted as they are very low ($<0.2$).}
\label{table:mae_results}
\begin{tabular}{c|c|llll|llll|l}
\multirow{3}{*}{\textbf{Dataset}}                                      & \multirow{3}{*}{\textbf{Model}} & \multicolumn{8}{c|}{\textbf{MAE (in days) for Time Prediction}}                                                                                                                                                                                                               & \multicolumn{1}{c}{\multirow{3}{*}{\textbf{\begin{tabular}[c]{@{}c@{}}Overall\\ MAE \\ (days)\end{tabular}}}} \\ \cline{3-10}
                                                                      &                                 & \multicolumn{4}{c|}{\textbf{Quartiles based on Events}}                                                                               & \multicolumn{4}{c|}{\textbf{Quarters based on Duration}}                                                                              & \multicolumn{1}{c}{}                                                                                          \\ \cline{3-10}
                                                                      &                                 & \multicolumn{1}{c|}{\textbf{1}} & \multicolumn{1}{c|}{\textbf{2}} & \multicolumn{1}{c|}{\textbf{3}} & \multicolumn{1}{c|}{\textbf{4}} & \multicolumn{1}{c|}{\textbf{1}} & \multicolumn{1}{c|}{\textbf{2}} & \multicolumn{1}{c|}{\textbf{3}} & \multicolumn{1}{c|}{\textbf{4}} & \multicolumn{1}{c}{}                                                                                          \\ \hline
\multirow{5}{*}{Helpdesk}                                              & GCN$_W$                            & 2.2955                          & \textbf{2.8397}                 & 4.1637                          & 0.3340                          & 3.6811                          & 6.4332                          & 3.6726                          & 0.1806                          & 2.3346                                                                                                        \\
                                                                      & GCN$_B$                            & 2.2993                          & 2.8577                          & 4.1483                          & \textbf{0.3143}                 & 3.6958                          & 6.3667                          & 3.4909                          & \textbf{0.1768}                 & 2.3298                                                                                                        \\
                                                                      & GCN$_{LB}$                           & 2.2973                          & 2.8474                          & 4.1085                          & 0.3433                          & 3.6744                          & 6.2572                          & 3.5081                          & 0.2020                          & 2.3250                                                                                                        \\
                                                                      & GCN$_{LW}$                           & 2.2950                          & 2.8470                          & \textbf{4.0661}                 & 0.3323                          & 3.6651                          & 6.1060                          & \textbf{3.2253}                 & 0.2195                          & \textbf{2.3095}                                                                                               \\
                                                                      & MLP                             & \textbf{2.2948}                 & 2.9030                          & 4.1969                          & 0.3445                          & \textbf{3.5724}                 & \textbf{5.688}                  & 5.0011                          & 0.3572                          & 2.3661                                                                                                        \\ \hline
\multirow{5}{*}{\begin{tabular}[c]{@{}c@{}}BPI'12 \\ (W)\end{tabular}} & GCN$_W$                            & \textbf{1.0956}                 & 1.5503                          & 1.6047                          & 1.1491                          & 1.7064                          & 2.4116                          & 1.7891                          & 0.4943                          & 1.3468                                                                                                        \\
                                                                      & GCN$_B$                            & 1.1134                          & 1.6109                          & 1.6877                          & 1.1449                          & 1.7666                          & 2.5344                          & 1.8986                          & 0.4548                          & 1.3837                                                                                                        \\
                                                                      & GCN$_{LB}$                           & 1.1114                          & 1.6043                          & 1.6775                          & 1.1359                          & 1.7530                          & 2.5318                          & 1.8997                          & \textbf{0.4495}                 & 1.3765                                                                                                        \\
                                                                      & GCN$_{LW}$                           & 1.1069                          & 1.5900                          & 1.6632                          & 1.1437                          & 1.7528                          & 2.4998                          & 1.8530                          & 0.4618                          & 1.3710                                                                                                        \\
                                                                      & MLP                             & 1.0966                          & \textbf{1.5224}                 & \textbf{1.5587}                 & \textbf{1.1288}                 & \textbf{1.6529}                 & \textbf{2.3617}                 & \textbf{1.7134}                 & 0.5276                          & \textbf{1.3229}                                                                                              
\end{tabular}
\end{table*}

\subsection{GCN Model Variants and MLP Baseline}
\label{subsec:model_variants}
We have introduced four GCN variants of this general architecture and an MLP-only variant for the experiments carried out in this work. 

\paragraph{GCN$_W$ (GCN with Weighted Adjacency Matrix)}
\label{subsub: weighted}
The adjacency matrix of the process graph depicted in Figure~\ref{fig:dfg} is computed. 
Rather than a traditional approach of using binary entries (as in $B_{BPI'12(W)}$), we introduce a new method in this variant by having the adjacency matrix store the values corresponding to the weighted edges of the process graph. 
The normalization procedure given in Eq.~\eqref{equation:d_normal} is then applied to this adjacency matrix in the GCN layer.

\paragraph{GCN$_B$ (GCN with Binary Adjacency Matrix)}
This variant uses the binary adjacency matrix shown in the previous section (see example: $B_{BPI'12(W)}$). 
The degree matrix is computed, from which a symmetrically normalized adjacency matrix is obtained. 
The main motivation behind using the binary and weighted variants of the adjacency matrix is due to the fact that GCN$_B$ is heavily influenced by outliers whereas GCN$_{W}$ might be biased by frequency differences between common connections in the DFG.

\paragraph{GCN$_{LW}$ (GCN with Laplacian Transform of Weighted Adjacency Matrix)}
The Laplacian matrix~\cite{godsil2013algebraic} of a graph is
$L = D - A$, where \textit{D} is the Degree matrix and the \textit{A} is the Adjacency matrix. In this variant, \textit{A} corresponds to the weighted adjacency matrix. The Laplacian matrix is then used for all computations involved within the Graph Convolutional layer as follows: $f(X,A,W) = \sigma (D^{-\frac{1}{2}}(D-A)D^{-\frac{1}{2}}XW)$.

\paragraph{GCN$_{LB}$ (GCN with Laplacian Transform of Binary Adjacency Matrix)}
This variant is equivalent to the previous one, except for the fact that it uses the binary adjacency matrix instead of the weighted adjacency matrix to compute the Laplacian matrix. 
\paragraph{MLP (Multi-Layer Perceptron)}:
In order to understand if the GCN layer added any significant change to the performance, we used a variant which had only the three fully-connected layers (omitting the GCN layer). 
This model also serves as baselines for the other architectures compared.
The feature matrix (\emph{X}) was flattened and given as input to the fully-connected layers. 
As in the other variants, Dropout is used before the last layer.
Hence, the dimensions for the input vector of the MLP was (\textit{$number\_of\_nodes \times number\_of\_features$}).

\subsection{Measures}

Each row is associated with two labels, the next activity type and the time (in seconds) after which the next event in that case takes place. 
As in \cite{tax2017predictive}, an additional label is added to denote the end of a case. 
\paragraph{Next Activity and Timestamp}
The quality of the next activity is measured in terms of the accuracy of predicting the correct label. 
In the case of timestamp prediction, we use Mean Absolute Error (MAE) calculated in days.

\paragraph{Quartiles based on Events}
We have evaluated the performance of each variant at different quartiles. 
The quartiles for each case instance have been computed based on the number of events.
For each case instance, its full list of events are split into four (approximately) equal quartiles, based on the order the events occurred in that case instance.  

\paragraph{Quarters based on Unit Length Time}
We normalize the full case duration to unit length time and divide it into four equidistant intervals, to make a comparison along the time axis between cases and datasets possible.
Thus, each case instance's full duration is divided by 4, and the case's events are put into the four intervals based on their individual finishing timestamps.
In contrast to the quartiles based on events above, these temporal quarters divide the true natural distribution of the process events based on time.

\section{Results of Predicting Event Types and Time at Different Stages for GCNs and MLP}
\label{sec:results}
\label{sec:q-results}

We describe per dataset the results for the GCN and MLP models based on the quartiles over event type
and quarters of the unit length time.
Subsequently, we compare the performance of the deep-learning architectures CNN, LSTM, GCN, and GAN with the MLP and classical approaches.

\subsection{Helpdesk Dataset}

\paragraph{Optimization}
Each of the model variants was initially run with different learning rates for the Adam optimizer.
The learning rate with the best performance was chosen for each variant. 
For all the GCN variants, the best performance for the timestamp predictor was obtained with a learning rate of 0.001. For the event predictor, GCN$_{LW}$ gave the best performance at a learning rate of 0.001 and all other GCN variants performed best at 0.0001. For the MLP model, both the tasks gave best results at a learning rate of 0.0001.
The model corresponding to the best validation loss is saved for all the model variants, and then evaluated on the same test set.

\paragraph{Results}
The accuracy values corresponding to the event-prediction task achieved in this process is presented in Table \ref{table:accuracy_results}. 
The Mean Absolute Error (in days) achieved on a test set from models saved for the different variants is shown in Table \ref{table:mae_results}.
It can be observed from Tables~\ref{table:accuracy_results} and \ref{table:mae_results} that the MLP model outperforms all other variants for the event-prediction task, in all individual quartiles/quarters as well as for the overall performance. Among all model variants, a maximum overall accuracy of 82.01\% is obtained for the event predictor by the MLP. The minimum overall MAE of 2.3095 days was achieved by the GCN$_{LW}$ variant. 

\subsection{BPI'12 (W) Dataset}
\paragraph{Optimization}
The same optimization procedure as for the Helpdesk dataset has been used.
The timestamp predictor for all variants gave the best results with a learning rate of 0.0001. 
It is also the preferred learning rate for the event predictor in all variants, except GCN$_{B}$ and MLP (where it is 0.00001). 
The computation of quartiles over event types
is also the same as before. 

\paragraph{Results}
The accuracy values and MAE values for the BPI'12 (W) dataset are presented in Tables~\ref{table:accuracy_results} and~\ref{table:mae_results}. 
The MLP model outperforms all other variants in the time-prediction task  for most of the scenarios. An overall minimum MAE of 1.3229 days is achieved. 
We are able to observe slight variations when it comes to the results of the event predictor. 
The best performance at individual quartiles and quarters are shown by GCN$_{LB}$ and MLP for different instances. The highest overall accuracy of 65.69\% is achieved by GCN$_{LB}$.

\begin{table*}[t!]
\centering
\caption{Comparison of the different models with other reported results on the same benchmark datasets}
  \begin{threeparttable}
\label{table:compare}
\begin{tabular}{l|l|l|l|l}
\multicolumn{1}{c|}{\multirow{2}{*}{\textbf{Model}}} & \multicolumn{2}{c|}{\textbf{\begin{tabular}[c]{@{}c@{}}Accuracy for \\ Event Prediction\end{tabular}}} & \multicolumn{2}{c}{\textbf{\begin{tabular}[c]{@{}c@{}}MAE (in days) for \\ Time Prediction\end{tabular}}} \\ \cline{2-5} 
\multicolumn{1}{c|}{}                                & \multicolumn{1}{c|}{\textit{Helpdesk}}            & \multicolumn{1}{c|}{\textit{BPI'12 (W)}}           & \multicolumn{1}{c|}{\textit{Helpdesk}}              & \multicolumn{1}{c}{\textit{BPI'12 (W)}}              \\ \hline
CNN~\cite{pasquadibisceglie2019using} & 0.7393 & \textbf{0.7817} & N/A & N/A \\
\hline
LSTM (Evermann et al.)~\cite{evermann2016deep} & N/A & 0.623 & N/A & N/A \\ 
LSTM (Camargo et al.)~\cite{camargo2019learning} & 0.789 & 0.778 & N/A & N/A \\
LSTM (Tax et al.)~\cite{tax2017predictive} & 0.7123 & 0.7600 & 3.75 & 1.56 \\ \hline

GCN$_W$ & 0.7954 & 0.6484 & 2.3346 & 1.3468 \\
GCN$_B$ & 0.7731 & 0.6473 & 2.3298 & 1.3837 \\
GCN$_{LB}$ & 0.7863 & 0.6569 & 2.3250 & 1.3765 \\
GCN$_{LW}$ & 0.7830 & 0.6525 & \textbf{2.3095} & 1.3710 \\
\hline
MLP & \textbf{0.8201} & 0.6559 & 2.3661 & \textbf{1.3229} \\ 

\hline
Breuker et al. \cite{breuker2016comprehensible} & N/A & 0.719 & N/A & N/A\\
WMP Van der Aalst et al. \cite{van2011time} & N/A  & N/A & 5.67 & 1.91 \\             
\hline
\hline
GAN+LSTM~\cite{taymouri2020predictive} ($k=2$) \tnote{a}& 0.8668 &  0.7535 &  1.6434 &  1.4004\\

GAN+LSTM~\cite{taymouri2020predictive}  ($k=4$) \tnote{a}& 0.8657 &  0.8009 &  1.1505 &  1.1611 \\
GAN+LSTM~\cite{taymouri2020predictive}  ($k=6$) \tnote{a} & 0.8976 &   0.8298 &  0.8864 &  0.9390 \\
GAN+LSTM~\cite{taymouri2020predictive} ($k=16$) \tnote{a} & N/A & 0.9019 & N/A &  0.4274 \\
GAN+LSTM~\cite{taymouri2020predictive} ($k=30$) \tnote{a} & N/A &  0.9290 & N/A &  0.3399 \\
\hline
LSTM (Lin et al.)~\cite{lin2019mm} \tnote{b} & 0.916  & N/A  & N/A & N/A \\
\hline
\end{tabular}
\begin{tablenotes}
\small{
\item a) Our reruns of the code adapted to fit the evaluation strategy of the CNN, LSTM, GCN, and MLP for fair comparison. 
Note, models are based on a specific $k$ value, i.\,e., they only predict cases of length $k+1$ or longer.
\item b) Code was not available. Thus the number cannot be independently confirmed.} \end{tablenotes}
\end{threeparttable}
\end{table*}

\section{Results of Comparing Deep-Learning Variants of CNN, LSTM, GCN, and GAN}
\label{sec:results_comparison}
We compare the performance of the different deep-learning variants of CNN, LSTM, GCN, and GAN with the MLP and classical approaches.
As mentioned in Section \ref{sec:rel_works}, the task of event prediction and the timestamp prediction has been explored in various other works as well, using other techniques. 
Table~\ref{table:compare} compiles the best results reported in other works and compares them with the results obtained from our GCNs as documented in Section~\ref{sec:q-results}. 

The values for the GAN by Taymouri et al.~\cite{taymouri2020predictive} have been obtained after rerunning the original code with necessary changes to make it comparable with the other results. 
This was necessary since the original paper by Taymouri et al.~\cite{taymouri2020predictive} reported only weighted average measures over different case lengths (\emph{k} values). 
Also, their train-test split ratio was 80:20 and changed to 66:33 as in the other and our models~\cite{tax2017predictive,pasquadibisceglie2019using}. 
The source code from the model introduced by Lin et al.~\cite{lin2019mm} was not available online. 
Hence, their results have been included in Table~\ref{table:compare} as a separate block. 
For the classical process-mining model reported by Van der Aalst et al. \cite{van2011time}, we have used the values obtained from the experiments conducted by Tax et al.~\cite{tax2017predictive} on the current datasets.


It can be observed from Table \ref{table:compare} that all the model variants introduced in this work perform well in comparison to previous models for the time-prediction task.
For the event-prediction task, we have mixed results. 
On the Helpdesk dataset, all the GCN model variants outperform two LSTM models~\cite{tax2017predictive,camargo2019learning} and the CNN model~\cite{pasquadibisceglie2019using}, but fail to outperform the improved LSTM model introduced by Lin et al. \cite{lin2019mm}. 
Our models perform poorly on the BPI'12 (W) dataset for event prediction.
Regarding the GAN+LSTM~\cite{taymouri2020predictive}, the results show that it is generally a strong performer.
But it has to be noted that the training procedure is fundamentally different from the other models due to the use of the parameter $k$.
This parameter denotes that subsequences of the processes of length $k$ are used for training, and $k+1$, $k+2$ etc. are used for testing. 
Thus, the result for, e.\,g., $k=30$ on the BPI12 (W) dataset only considers few process cases of length 31 or more.

\section{Discussion}
\label{sec:discussions}

Our experiments show that a simple MLP is able to outperform other sophisticated architectures such as the LSTMs and CNN. 
But it is also to be noted that MLP does not emerge as the best performer in all of the experiments. 
Some possible factors that might have resulted in this performance could be an improved feature vector representation or the fact that the number of classes in the event-prediction task is not that high (9+1 classes for Helpdesk dataset and 6+1 classes for the BPI'12 (W) dataset). 
Thus, the simple MLP models were able to effectively learn the correlations between input features and the target labels. 

Regarding our analysis at different quartiles based on the number of events
and quarters based on unit-length time 
show that automated process-prediction results vary at different stages of a business process.
For example, with the Helpdesk dataset, the accuracy of event prediction continuously improves over the quartiles 
based on events.
However, for the BPI'12 (W) dataset, it surprisingly improves only until the 3rd quartile, when it suddenly drops in the last quartile.
A similar observation can be made for MAE 
over both quartiles based on events and quarters based on duration. 
Here, the scores continuously increase (MAE gets worse), until they drop in the last quartile.
Quartiles over events and quarters over unit length time truly model two different things. Quarters better reflect the performance in a unit length progression over time, but can be negatively influenced by a skewed event distribution. At the same time, quartiles have an equal distribution.
Future experiments would need to be conducted to explain this varying behaviour between datasets and measures.

A potential risk to the validity of these results can be from one of the assumptions we had used during the pre-processing stage. 
Where there were recurring events of the same type in a case, we only included that event type's most-recent occurrence. 
Particularly in the BPI'12 (W) dataset, there are cases where the same event occurs many times. 
To understand how our assumption might have affected the results, the same experiments were performed on a different version of the BPI'12(W) dataset, which had reduced instances of an event following itself~\cite{tax2017predictive}. 
But the results obtained were very similar to the original dataset.

Comparing the different models has been in general very difficult, due to different train-test split ratios and different training procedures. 
Following~\cite{tax2017predictive,pasquadibisceglie2019using}, we have used $2/3${rd} of the data for training and $1/3$rd for testing, while preserving the chronological nature of the data. 
Other works like \cite{breuker2016comprehensible,camargo2019learning} have also used a ratio that is comparable to ours, namely 70:30 for training and testing. 
Only the GAN model \cite{taymouri2020predictive} had originally used a 80:20 split and the work carried out by Lin et al.~\cite{lin2019mm} have split the data in a 7:2:1 ratio. 
Since the GAN code is available, we adapted it to the same train-test split and rerun it with $25$ epochs, as stated in the paper, for different values of $k$.
The code for the LSTM by Lin et al. is not available, as also noted by Taymouri et al.~\cite{taymouri2020predictive}, and thus cannot be independently confirmed.
However, this study includes three other strong LSTM models, which are directly comparable.

A key difference of the GAN model is its training procedure, which involves windows of different case-lengths (the \emph{k} values), whereas our training procedure does not differentiate between different case lengths. 
For example, the GAN model with $k=30$ is trained on subsequences of processes of a length of 30 in the BPI12 (W) dataset.
For testing, only the remaining few process cases of length 31, 32 etc. are used.
Thus, the GAN results ~\cite{taymouri2020predictive} cannot be compared directly to any of the other models, which are designed to make predictions on any lengths of cases, but are reported in Table~\ref{table:compare} for completeness.

The major impact of this work lies in the observation that there is no silver-bullet method when it comes to business-process prediction. It can be observed that MLP is a strong baseline and in many cases outperforms complex neural networks like the LSTM, GCN, and CNN. 
However, interestingly, there are cases where the MLP performs comparably poor, such as predicting the activity type in the BPI2 (W) dataset.
There have been other works which report similar behaviour of an MLP baseline for classification tasks~\cite{DBLP:conf/um/GalkeMVS18,IJCNN-GalkeEtAl-2021,DBLP:conf/jcdl/MaiGS18}. 
Thus, interesting future work is to understand why MLPs perform well on certain datasets, outperforming strong models, while their performance is low for other datasets. 
Also, it would be interesting to look into other variations in representing the feature vector.

\section{Conclusions}
\label{sec:conclusions}

Our experiments show that MLP is a strong baseline for the task of event prediction and time prediction in business processes. 
However, overall the MLP is not a clear best performing model.
Furthermore, the detailed analyses at different quartiles
based on the number of events and quarters based on unit length time 
show that automated process-prediction results vary at different stages of a business process. 
Hence, care must be taken while evaluating and applying business-process prediction models. 
The source code for 
this work is available at: 
\url{https://github.com/ishwarvenugopal/GCN-ProcessPrediction}

\begingroup
\setstretch{0.9}
\setlength\bibitemsep{0pt}
\printbibliography
\endgroup

\end{document}